\documentclass[runningheads]{llncs}
\usepackage[parfill]{parskip}
\usepackage[margin=4cm]{geometry}
\usepackage{orcidlink}
\usepackage[misc,geometry]{ifsym}
\usepackage[square,sort,comma,numbers]{natbib}

\begin{document}
\title{Genesis: Towards the Automation of Systems Biology Research}

\author{Ievgeniia A. Tiukova\inst{1,2}\orcidlink{0000-0002-0408-3515} \and
Daniel Brunnsåker\inst{1}\orcidlink{0000-0002-5167-0536} \and
Erik Y. Bjurström\inst{1}\orcidlink{0000-0002-8863-1207} \and
Alexander H. Gower\inst{1}\orcidlink{0000-0002-8358-0842} \and
Filip Kronström\inst{1}\orcidlink{0000-0002-3011-5541} \and
Gabriel K. Reder\inst{3}\orcidlink{0000-0001-8918-0789} \and
Ronald S. Reiserer\inst{4}\orcidlink{0000-0002-3786-7893} \and
Konstantin Korovin\inst{5}\orcidlink{0000-0002-0740-621X} \and
Larisa B. Soldatova\inst{6}\orcidlink{0000-0001-6489-3029} \and
John P. Wikswo\inst{4}\orcidlink{0000-0003-2790-1530} \and
Ross D. King\inst{1,3}\orcidlink{0000-0001-7208-4387}\Letter}

\authorrunning{King \emph{et al.}}
\institute{Chalmers University of Technology, Gothenburg, Sweden
\email{\{tiukova,danbru,erikbju,gower,filipkro,rossk\}@chalmers.se}
\and KTH Royal Institute of Technology, Stockholm, Sweden
\and University of Cambridge, Cambridge, United Kingdom
\email{gr513@cam.ac.uk}
\and Vanderbilt University, Nashville, TN, USA
\email{\{ron.reiserer,john.p.wikswo\}@vanderbilt.edu}
\and The University of Manchester, Manchester, United Kingdom
\email{konstantin.korovin@manchester.ac.uk}
\and Goldsmiths, University of London, London, United Kingdom
\email{l.soldatova@gold.ac.uk}} 

\maketitle
\begin{abstract}
    The cutting edge of applying AI to science is the closed-loop automation of scientific research: robot scientists. We have previously developed two robot scientists: `Adam' (for yeast functional biology), and `Eve' (for early-stage drug design)). We are now developing a next generation robot scientist Genesis. With Genesis we aim to demonstrate that an area of science can be investigated using robot scientists unambiguously faster, and at lower cost, than with human scientists. Here we report progress on the Genesis project. Genesis is designed to automatically improve system biology models with thousands of interacting causal components. When complete Genesis will be able to initiate and execute in parallel one thousand hypothesis-led closed-loop cycles of experiment per-day. Here we describe the core Genesis hardware: the one thousand computer-controlled $\mu$-bioreactors. For the integrated Mass Spectrometry platform we have developed AutonoMS, a system to automatically run, process, and analyse high-throughput experiments. We have also developed Genesis-DB, a database system designed to enable software agents access to large quantities of structured domain information. We have developed RIMBO (Revisions for Improvements of Models in Biology Ontology) to describe the planned hundreds of thousands of changes to the models. We have demonstrated the utility of this infrastructure by developed two relational learning bioinformatic projects. Finally, we describe LGEM+ a relational learning system for the automated abductive improvement of genome-scale metabolic models.

\end{abstract}

\section{Background}
\subsection{Automating Science}
The most advanced application of AI for science is the closed-loop automation of scientific research. Such systems are called `Robot Scientists', `AI Scientists', `Self-driving Labs', etc. A Robot Scientist autonomously originates hypotheses to explain observations, devise experiments to test these hypotheses, physically runs the experiments using laboratory robotics, interprets the results to change the probability of hypotheses, and then repeats the cycle \cite{king_functional_2004,king_automation_2009}. The automation of science has the potential to revolutionise the efficiency of scientific research \cite{king_artificial_2023,king_framework_2023,king_robot_2023}.

We have previously developed two Robot Scientists, `Adam', and `Eve'.  Adam was the first machine to autonomously discover novel scientific knowledge \cite{king_automation_2009}. Adam investigated the functional genomics of the yeast \emph{S. cerevisiae}, and discovered the function of locally orphan enzymes.  Our second Robot Scientist, Eve, was designed for early-stage drug development \cite{williams_cheaper_2015}. Using econometric modelling we demonstrated that Eve's use of AI to select compounds outperformed standard drug screening. Eve's most significant discovery is that triclosan (an anti-microbial compound commonly used in toothpastes, etc.) is an inhibitor of wild-type and drug-resistant dihydrofolate reductase in the malaria-causing parasites \emph{P. falciparum} and \emph{P. vixax} \cite{bilsland_yeast-based_2013,williams_cheaper_2015,bilsland_plasmodium_2018}. Eve's approach for closed-loop optimisation has been widely copied by the pharmaceutical industry and in materials science.

\subsection{Eukaryotic System Biology}
One of the most challenging tasks in modern science is the development of systems biology models of eukaryotic cells. These models are central to the future of medicine (humans are eukaryotes), to agriculture (plants are also eukaryotes), and to biotechnology. The yeast \emph{S. cerevisiae} is the `model organism' for eukaryotic cells: despite the last common ancestor of humans and yeast living around one billion years ago, most of what is true about yeast is also true for human cells. The medicine Nobel Laureate Leland Hartwell famously said `yeast are like small humans'. The similarities between yeast and humans mean that often the easiest way to understand how a human gene works is to study its homolog in yeast, as yeast is much easier to work with.

Even simple model biological system like yeast are incredibly complicated: thousands of genes, proteins, and small-molecules, all interacting together in complicated spatial temporal ways.  In addition, Ockham's razor is not a reliable guide in biology, as biological systems evolved over long time periods.  Therefore, basic information theory makes clear that a vast number of experiments will be required to derive an understanding of biological systems. Current high-throughput methods are insufficient for systems biology.  This is because, even though very large numbers of experiments may be executed, each individual experiment cannot be designed to test a hypothesis about a model, i.e. current high-throughput experiments are not `hypothesis led'.  

Systems biology presents an extreme challenge to the traditional human based scientific method. Systems biology models are so complex that they are beyond human intuitive understanding. This complexity plays to the strength of AI. Complex modelling is also complicated by what is known in the philosophy of science as `Duhem Thesis': `an experiment … can never condemn an isolated hypothesis but only a whole theoretical Group' \cite{duhem_aim_1991}.  This makes model refinement and the generation of efficient experiments to test models especially challenging \cite{kitano_nobel_2021}. We argue that AI systems are now better than humans at this. Due to these, and other challenges of systems biology, AI tools are required to aid the execution of the many closed-loop experimental cycles essential to build accurate and comprehensive 21st century system biology models.

As a proof-of-principle example of the automation of systems biology, we previously chose the diauxic shift in the yeast \emph{S. cerevisiae} \cite{coutant_closed-loop_2019}. We automatically developed a model that outperformed the best previous models, and added 92 genes (+45\%), and 1,048 interactions (+147\%) \cite{coutant_closed-loop_2019}. The resulting improved model is relevant to understanding cancer, the immune system, and aging.

\section{Results}
\begin{figure}
    \centering
    \includegraphics[width=\textwidth]{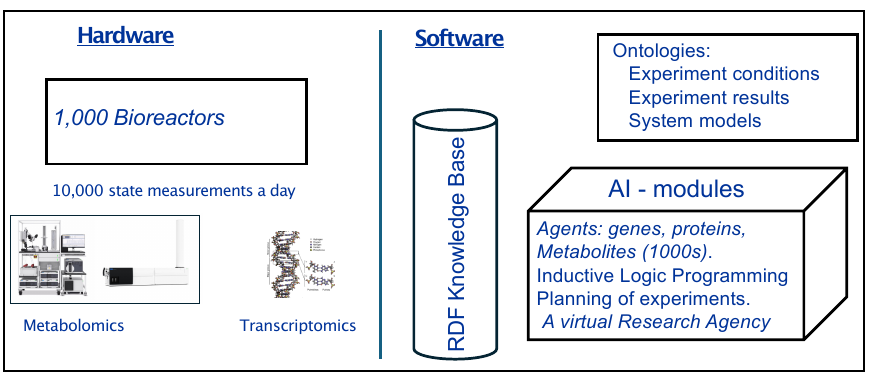}
    \caption{The overall architecture of the Genesis system. The main hardware system will have 1,000 computer-controlled $\mu$-bioreactors. These will be connected to a mass-spectrometer to read-out the metabolomic (small-molecules) state of the yeast population, and to an RNA-SEQ system to read-out the transcriptome (tRNA) state of the yeast population. The software consists of many modules, going from low-level bioreactor control, to high-level AI units. The units in italics are currently incomplete.}
    \label{fig:gen}
\end{figure}

The overall architecture of Genesis is shown in Fig.~\ref{fig:gen}. With Genesis we aim to demonstrate that an area of science can be investigated unambiguously faster, and at lower cost, using robot scientists than with human scientists. Specifically, we aim to demonstrate a hundred-fold cost-benefit for Genesis over a standard human scientist in a standard lab.

Genesis is designed to automatically improve a more complicated scientific model than any previous system has attempted. The model that Genesis  works with has thousands of interacting causal (mechanical) components related by tens of thousands of parameters. This contrasts sharply with all other closed-loop automated systems that we are aware of, which focus on simple input/output black-box optimisation.

Genesis is specifically designed to automate systems biology research. Genesis will be designed to be able to initiate and execute in parallel one thousand hypothesis-led closed-loop cycles of experiment per-day. Each closed-loop cycle will consist of hypothesis formation, experiment planning, laboratory execution, and results interpretation.

\subsection{Genesis Hardware}
\begin{figure}
    \centering
    \includegraphics[width=\textwidth]{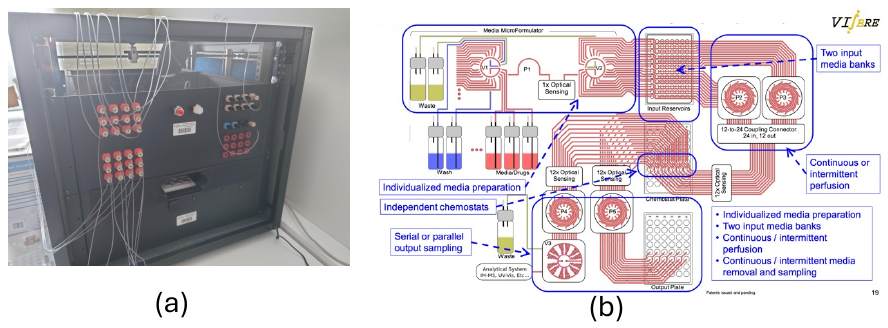}
    \caption{Genesis Hardware. (a) Is the initial 12 $\mu$-bioreactor system working in Chalmers. (2) The schematics of the fluidics and micro-formulator design.}
    \label{fig:genesis_hw}
\end{figure}

The hardware of Genesis will have at its heart a micro-fluidic system with one thousand computer-controlled $\mu$-bioreactors (developed in Vanderbilt University, USA) (Fig.~\ref{fig:genesis_hw}). Achieving this will be a revolution in laboratory automation, as most biological labs have $<$10 chemostats. The $\mu$-bioreactors will be arranged in groups of 48 using the standard footprint of a microtitre plate. Each $\mu$-bioreactors will be able to be configured in real time to run in batch, fed batch, or continuous mode. This flexibility will enable a very wide range of biological conditions to be explored, and experiments to be executed. In each $\mu$-bioreactors the Genesis-AI system will specify experiments of the form:

\begin{itemize}
    \item Genetics: (yeast strain - from a large library kept in an automated deepfreeze: $\sim$20,000 strains deletants, reporters, etc.), growth-rate (when used as a chemostat), OD (when used as a turbidostat).
    \item Environment: Growth medium (a cocktail of 10 metabolites and small-molecules added to a minimally defined medium from $\sim$100), Drugs (a complex cocktail of compounds added to the growth media from $\sim$10,000). 
\end{itemize}

The observables from these experiments will be growth rate (batch), metabolic analysis of the growth medium ($\sim$10 compounds), metabolic analysis of the internal state of the yeast ($\sim$100 metabolites), and comprehensive gene expression levels ($\sim$6,000 genes). These observables are both automatable and highly informative.

\subsection{Mass Spectrometry}
In our proof of concept work on automating systems biology \cite{coutant_closed-loop_2019} we could only measure yeast population growth - using Optical Density (OD). This greatly limited how much could be learnt about systems biology. With Genesis we will measure (1) the internal state of $\sim$100 metabolites in the yeast populations in the $\mu$-bioreactors, and (2) the internal state of $\sim$6,000 tRNA types in the yeast populations in the $\mu$-bioreactors (RNA-SEQ). This will enable far greater information constraints on the modelling.

\begin{figure}
    \centering
    \includegraphics[width=\textwidth]{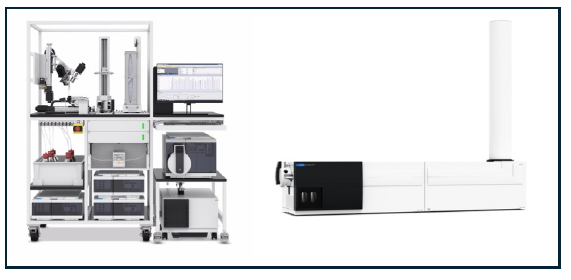}
    \caption{The Gensis Mass Spectrometry platform: an Agilent RapidFire (the robotics) and a 6560  ion mobility-mass spectrometry (IM-MS) system.}
    \label{fig:ms}
\end{figure}

For the metabolomics experiments we plan a capacity of $\sim$10,000 measurements a day. This is possible through integrating laboratory automation (Agilent RapidFire) and an Agilent 6560  ion mobility-mass spectrometry (IM-MS) system (Fig.~\ref{fig:ms}). Ion mobility MS does not require a slow chromatography step. To the best of our knowledge this is a higher rate of measurements than any existing MS system. To deal with this planned high rate of measurement we have developed AutonoMS, a platform for automatically running, processing, and analysing high-throughput mass spectrometry experiments \cite{reder_autonoms_2024}. AutonoMS enables automated software agent-controlled end-to-end measurement and analysis runs from experimental specification files that can be produced by human users or upstream software processes. AutonoMS is currently designed for IM-MS, but can be adapted to additional analytical instruments and data processing flows.

We have demonstrated the utility of AutonoMS in a high-throughput flow-injection ion mobility configuration with 5 second sample analysis time, processing robotically-prepared chemical standards and cultured yeast samples in targeted and untargeted metabolomics applications \cite{reder_autonoms_2024}.

\subsection{Ontologies}
In collaboration with IT company Thoughtworks, we have developed Genesis-DB, a database system designed to support the Genesis project, and Robot Scientists in general, by providing software agents access to large quantities of structured domain (RDF, Datalog) information \cite{reder_genesis-db_2023}. We have also developed a new ontology for modelling data and meta-data from autonomously performed yeast $\mu$-bioreactors cultivations. The ontologies for experiments records such experimental conditions as temperature, growth medium, pH, sampling times, etc. The ontology for experimental results records Gene counts, mass spec readouts, etc. Genesis-DB is designed to support reasoning about past experiments, and the planning of  future experiments \cite{reder_genesis-db_2023}.

\begin{figure}
    \centering
    \includegraphics[width=0.9\textwidth]{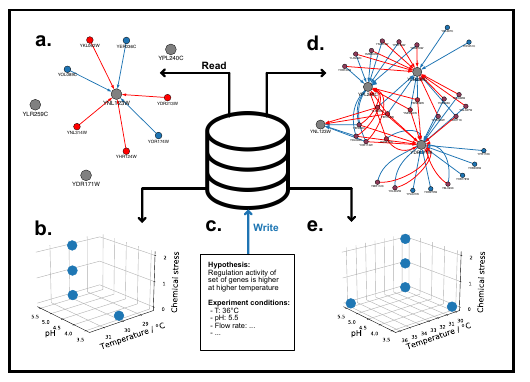}
    \caption{Visualisation of database usage from demonstration software agent utilisation of experimental metadata for systems biology model improvement. It represents a cycle of model improvement. (a) First a gene regulatory network is reconstructed from gene counts retrieved from the database with query. (b) Then the experimental conditions are retrieved and the space visualised. (c) A hypothesis along with the experimental procedures to test it are written to the database. (d) Then the regulatory network is recreated including data including the new high temperature experiment. (e) New visualisation with high temperature data.}
    \label{fig:gen_db}
\end{figure}

We have demonstrated how Genesis-DB enables the research life cycle by modelling yeast gene regulation, guiding future hypotheses generation and design of experiments (Fig.~\ref{fig:gen_db}) \cite{reder_genesis-db_2023}. Genesis-DB supports AI-driven discovery through automated reasoning and its design is portable, generic, and easily extensible to other AI-driven molecular biology laboratory data and beyond. 

\subsection{Model Revision}
\begin{figure}
    \centering
    \includegraphics[width=\textwidth]{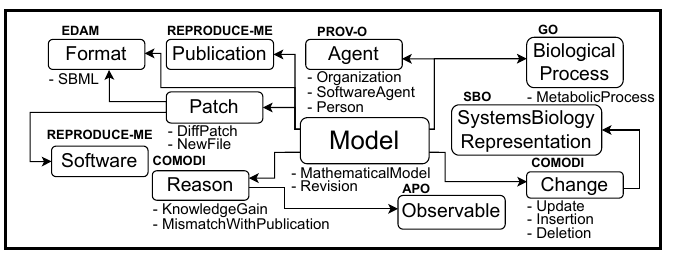}
    \caption{Overview of RIMBO, showing the classes and their relationships.}
    \label{fig:rimbo}
\end{figure}

In the Genesis project we will automate hundreds of thousands of system biology model revisions. This implies that these model revisions will need to be systematically recorded and available for inference. We have therefore developed RIMBO (Revisions for Improvements of Models in Biology Ontology), which describes the changes made to computational biology models \cite{kronstrom_rimbo_2023} (Fig.~\ref{fig:rimbo}). RIMBO is intended as the foundation of a database containing and describing iterative improvements to models. By recording high level information, such as modelled phenomena, and model type, using controlled vocabularies from widely used ontologies, the same database can be used for different model types. The database aims to describe the evolution of models by recording chains of changes to them. To make model revisions clear, emphasise has been put on recording the reasons, and descriptions, of the changes.

We have demonstrated the usefulness of a database based on RIMBO by modelling the update from version 8.4.1 to 8.4.2 of the genome-scale metabolic model Yeast8, a modification proposed by an abduction algorithm, as well as thousands simulated revisions \cite{kronstrom_rimbo_2023}. This results in a database demonstrating that revisions can successfully be modelled in a semantically meaningful and storage efficient way. 

\subsection{Bioinformatics}
To demonstrate the utility of our ontologies and databases we have used them as the foundation of two bioinformatic studies \cite{brunnsaker_high-throughput_2023,brunnsaker_interpreting_2024}. Even though \emph{S. cerevisiae} is a very well-studied organism, $\sim$20\% of its proteins remain largely unannotated. Many of these uncharacterized proteins are conserved between eukaryotes, including humans, providing a significant incentive to increase the pace of discovery.

The first bioinformatics investigation made use of untargeted metabolomics as a tool for functional discovery, generating profiles of ten regulatory deletant strains which are investigated to better understand the consequences of their deletion, and their role in the metabolic reconfiguration of the diauxic shift \cite{brunnsaker_high-throughput_2023}. Using previous semi-autonomously developed gene regulatory models produced by Eve \cite{coutant_closed-loop_2019}, a set of ten regulatory genes were selected due to both their particular relevance to the shift and implications of previously unknown connections in literature. These were then individually and collectively investigated using their untargeted metabolic profiles with the goal of clarifying regulatory roles and biological consequences of gene deletion. This also served as an assessment of the suitability of untargeted metabolomics as a tool for guidance in high-throughput model improvements.

The second bioinformatics investigation focussed on proteomics (the state of the proteins in the yeast population). Proteomic profiles reflect the functional readout of the physiological state of an organism. \emph{S. cerevisiae} is a well-studied model organism, and there is a large amount of structured knowledge on yeast systems biology in databases such as the Saccharomyces Genome Database, and highly curated genome-scale metabolic models like Yeast8. These data-sets, the result of decades of experiments, are abundant in information, and adhere to semantically meaningful ontologies \cite{brunnsaker_interpreting_2024}. By representing this knowledge in an expressive Datalog knowledgebase we generated data descriptors using relational learning that, when combined with supervised machine learning, enables us to predict protein abundances in an explainable manner. We learnt predictive relationships between protein abundances, function and phenotype; such as $\alpha$-amino acid accumulations and deviations in chronological lifespan. We further demonstrate the power of this methodology on the proteins His4 and Ilv2, connecting qualitative biological concepts to quantified abundances. 

\begin{figure}
    \centering
    \includegraphics[width=0.5\textwidth]{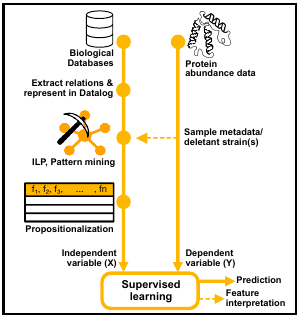}
    \caption{Data-set construction using frequent pattern mining on databases on yeast systems biology (SGD, Yeast8, KEGG, BioGRID) and sample metadata from measured protein abundances. Biological databases are represented in Datalog, patterns are then mined with the WARMR relational learning method (using information on strains as examples) (Dehaspe et al., 1998; Srinivasan, 2024) to extract propositional relational patterns (and simple instantiations) that are then used as independent variables in the prediction of protein abundances.}
    \label{fig:rel_learn}
\end{figure}

\subsection{Learning Logical Models}
The primary systems biology target for Genesis is the automated improvement of genome-scale metabolic models (GEMs) in \emph{S. cerevisiae}. There are many different approaches to the systems biology modelling of metabolism, ranging from ordinary differential equations (ODEs), flux-balance analysis (FBA) simulations, to graph topology models \cite{coutant_closed-loop_2019}. One approach we favour is to use first-order logic (FOL) \cite{king_functional_2004}. The advantage of using FOL is that there are many available tools to infer new structure; which is much harder to do with say ODEs. We have developed LGEM+, a system for automated abductive improvement of GEMs consisting of: a compartmentalised FOL framework for describing biochemical pathways (using curated GEMs as the expert knowledge source); and a two-stage hypothesis abduction procedure \cite{gower_lgem_2023}.

\begin{figure}
    \centering
    \includegraphics[width=0.7\textwidth]{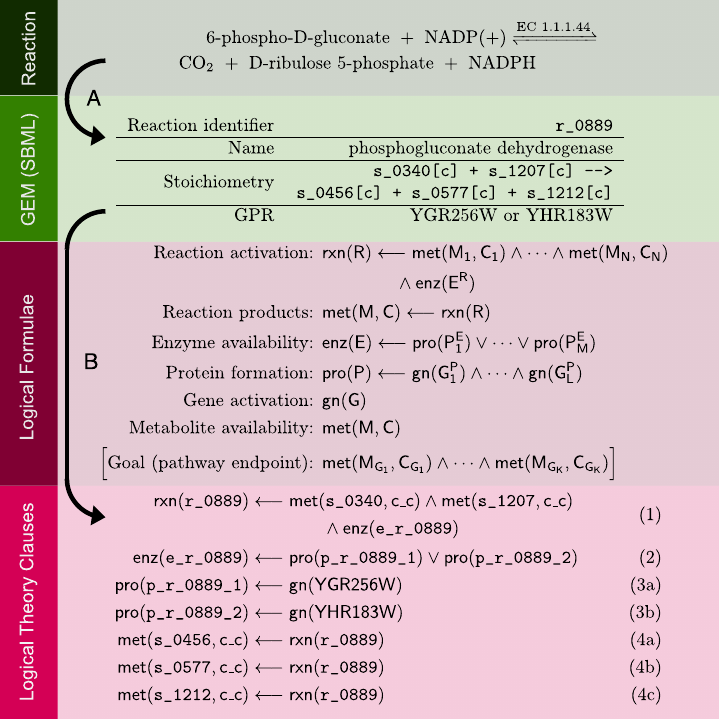}
    \caption{Conversion of genome-scale metabolic model provided in SBML to logical theory. (A) A reaction is encoded in SBML using identifiers to represent the substrates and products, and a logical rule for enzyme presence. (B) The information contained on each reaction is encoded using a logical formulae into a set of clauses. Here equation (1) is the reaction activation clause. it can be read as: “reaction r0889 is active if all of the metabolites in the set \{s0340, s1207\} are present in the cytoplasm and at least one of the isoenzymes is present”. Similarly, equation (2) describes the condition for a relevant enzyme to be present; equations (3a,b) describe the conditions for each of these isoenzymes to be formed; and equations (4a-d) are the reaction product clauses and state that ``if reaction r0889 is then each of its products are present''.}
    \label{fig:lgem}
\end{figure}

We have demonstrated that deductive inference on logical theories created using LGEM+, using the automated theorem prover iProver \cite{korovin_iprover_2008}, can predict growth/no- growth of S. cerevisiae strains in minimal media. LGEM+ proposed 2094 unique candidate hypotheses for model improvement \cite{gower_lgem_2023}. We assessed the value of the generated hypotheses using two criteria: (a) genome-wide single-gene essentiality prediction, and (b) constraint of flux-balance analysis (FBA) simulations. For (b) we developed an algorithm to integrate FBA with the logic model. We demonstrated a model-driven experimental design strategy, and demonstrate this with a differential expression study on the gene PFK2 \cite{gower_lgem_2023}.

\section{Discussion}
\subsection{Completing Genesis}
The core structure of Genesis is complete, but much remains before the system is working at full capacity (Fig.~\ref{fig:gen}). For the hardware: the $\mu$-bioreactor hardware needs to be scaled-up from 12 $\mu$-bioreactor units to 48 $\mu$-bioreactor units, and then 21 such units integrated; the MS system needs to fully integrated with the $\mu$-bioreactors; and we also need to integrate the planned (RNA-SEQ) transcriptomics experiments. For the software: we need more work on ontologies, databases, and knowledgebases; we need to develop the planned agent system (an agent for every known gene, protein, small molecule in yeast); we need to further develop the relational learning/inductive logic programming machine learning systems to enable the formation of more biologically relevant hypotheses about mode structure; we need to develop a robust control system to deal with the order of a thousand hypothesis-led experimental cycles per day; and finally we need to implement a Genesis virtual Research Agency to control overall resources to the agents.

\subsection{Large Language Models}
The success of  Large Language Models (LLMs) is a step-change in AI. LLMs have achieved breakthrough performance on a wide range of tasks that require human intelligence. It is to be expected that LLMs will impact on the Genesis project in multiple ways, but it is not yet clear which will be the most important. One exciting potential use of LLMs in Genesis is in converting scientific knowledge currently encoded in natural language (e.g. in textbooks, papers), into explicit formal knowledge (e.g. in logic) that can be checked for correctness/truth and can be directly reasoned with. LLMs solve the predicate invention problem.

Another exciting use of LLMs is in hypothesis formation. The architecture of LLMs entails that the output string is the most likely one given the input string and the training data. For science these strings may be interpreted as scientific hypotheses. Due to their internal complexity and sophistication LLMs have the potential to go beyond existing text-based scientific hypothesis generation tools. The generation of hypotheses by LLMs is closely related to the phenomena of ``hallucinations''. These are LLM outputs that are not valid inferences from the training data. Some hallucinations may be simply factually wrong. For others, their validity is uncertain. Hallucinations are a serious problem in many applications. For example, in science it is not acceptable to hallucinate (make up) false references. However in scientific hypothesis generation hallucinations may be useful: probable novel hypotheses whose validity may be objectively tested by laboratory experiments. 

\section*{Acknowledgments}
Funding: This work has been supported by the UK Engineering and Physical Sciences Research Council (EPSRC) [EP/R022925/2, EP/W004801/1 and EP/X032418/1], and by the Wallenberg AI, Autonomous Systems and Software Program (WASP) funded by the Alice Wallenberg Foundation.

\bibliographystyle{ieeetr}
\bibliography{references}
\end{document}